\documentclass{article}

\PassOptionsToPackage{numbers, compress}{natbib}
 \usepackage[preprint]{neurips_2026}

\usepackage[utf8]{inputenc} 
\usepackage[T1]{fontenc}    
\usepackage{hyperref}       
\usepackage{bookmark}
\usepackage{url}            
\usepackage{booktabs}       
\usepackage{amsfonts}       
\usepackage{amsmath}        
\usepackage{amssymb}        
\usepackage{nicefrac}       
\usepackage{microtype}      
\usepackage[dvipsnames,table]{xcolor}   
\usepackage{graphicx}       
\usepackage{subfig}         
\usepackage{multirow}       
\usepackage{colortbl}       
\usepackage{wrapfig}        
\usepackage[capitalize]{cleveref}  
\usepackage{algorithm}      
\usepackage{listings}       
\usepackage{tcolorbox}      

\makeatletter
\def\shline{\noalign{\ifnum0=`}\fi\hrule \@height 1pt \futurelet\reserved@a\@xhline}
\def\midline{\noalign{\ifnum0=`}\fi\hrule \@height 0.5pt \futurelet\reserved@a\@xhline}

\makeatother

\makeatletter
\renewcommand{\@bottomtitlebar}{
  \vskip 0.29in
  \vskip -\parskip
  \hrule height 1\p@
  \vskip 0.02in%
}
\renewcommand{\@maketitle}{%
  \vbox{%
    \hsize\textwidth
    \linewidth\hsize
    \vskip 0.1in
    \@toptitlebar
    \centering
    {\LARGE\bf \@title\par}
    \@bottomtitlebar
    \if@anonymous
      \begin{tabular}[t]{c}\bf\rule{\z@}{10\p@}
        Anonymous Author(s) \\
        Affiliation \\
        Address \\
        \texttt{email} \\
      \end{tabular}%
    \else
      \def\And{%
        \end{tabular}\hfil\linebreak[0]\hfil%
        \begin{tabular}[t]{c}\bf\rule{\z@}{10\p@}\ignorespaces%
      }
      \def\AND{%
        \end{tabular}\hfil\linebreak[4]\hfil%
        \begin{tabular}[t]{c}\bf\rule{\z@}{10\p@}\ignorespaces%
      }
      \begin{tabular}[t]{c}\bf\rule{\z@}{10\p@}\@author\end{tabular}%
    \fi
    \vskip 0.3in \@minus 0.1in
  }
}
\makeatother

\title{PixelGen: Improving Pixel Diffusion with \\Perceptual Supervision}

\author{%
  \textbf{Zehong~Ma$^{1}$,\quad Ruihan Xu$^{1}$,\quad Shiliang Zhang$^{1}$}\\
  $^{1}$State Key Laboratory of Multimedia Information Processing, \\ School of Computer Science, Peking University \\
}

\begin{document}

\maketitle

\begin{figure}[h]
    \centering
    \vspace{-2.5em}
    \includegraphics[width=\linewidth]{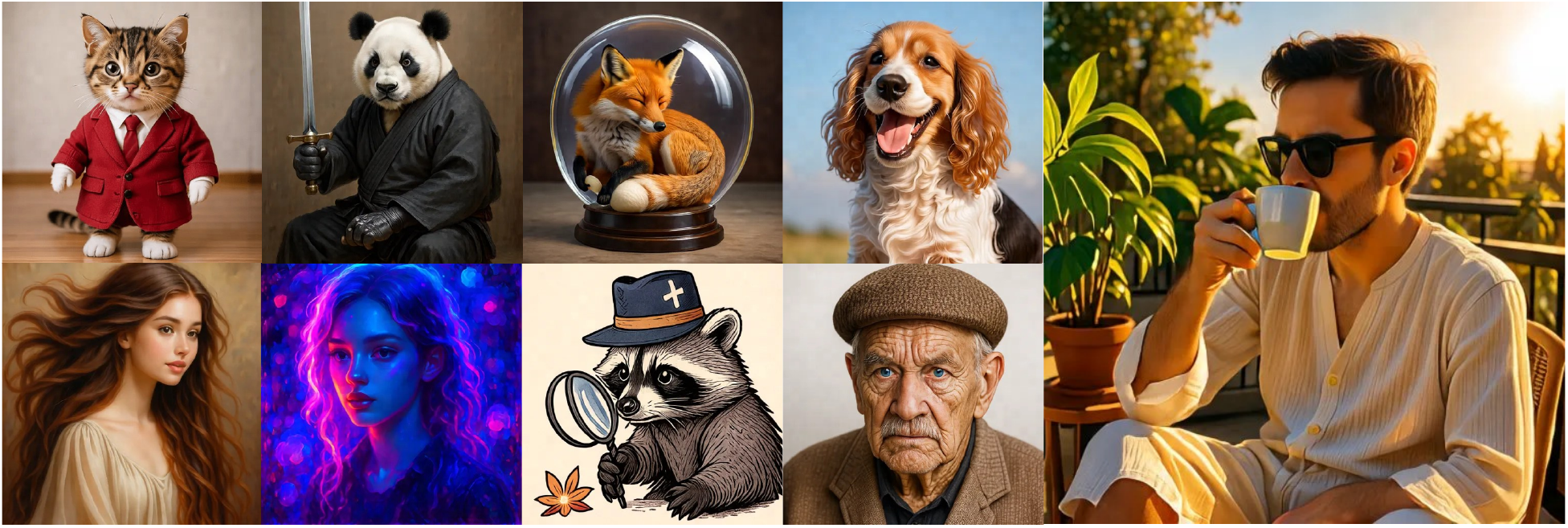}
    \caption{Text-to-image samples from PixelGen, pretrained with only 6 days on 8$\times$H800 GPUs.}
    \label{fig:t2i_visualization}
\end{figure}

\begin{abstract}
  Pixel diffusion generates images directly in pixel space, avoiding the VAE artifacts and representational bottlenecks of two-stage latent diffusion. Recent JiT further simplifies pixel diffusion with $x$-prediction, where the model predicts clean images rather than velocity. However, the standard pixel-wise diffusion loss treats all pixels equally, spending model capacity to perceptually insignificant signals and often leading to blurry samples. We propose PixelGen, an end-to-end pixel diffusion framework that augments $x$-prediction with perceptual supervision. Specifically, PixelGen introduces two complementary perceptual losses on top of $x$-prediction: an LPIPS loss for local textures and a P-DINO loss for global semantics. To preserve sample coverage, PixelGen further proposes a noise-gating strategy that applies these losses only at lower-noise timesteps. On ImageNet-256 without classifier-free guidance, PixelGen achieves an FID of 5.11 in 80 training epochs, surpassing the latent diffusion baselines. Moreover, PixelGen scales efficiently to text-to-image generation, reaching a GenEval score of 0.79 with only 6 days of training on 8$\times$H800 GPUs. These results show that perceptual supervision substantially narrows the gap between pixel and latent diffusion while preserving a simple one-stage pipeline. Codes are available at \url{https://github.com/Zehong-Ma/PixelGen}.
\end{abstract}

\section{Introduction}
\label{sec:intro}

\begin{figure}[t]
    \centering
    \includegraphics[width=0.95\linewidth]{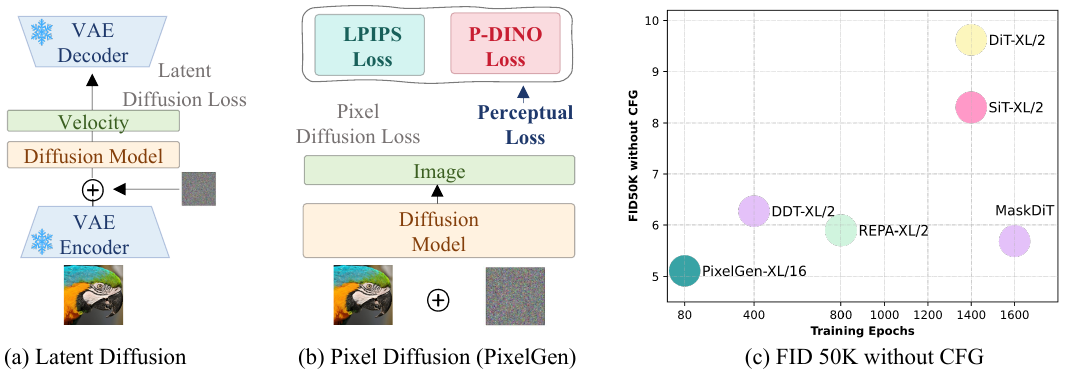}
    \caption{Perceptual supervision improves pixel diffusion. (a) Latent diffusion is affected by VAE bottlenecks. (b) Standard pixel diffusion relies on uniform pixel-wise supervision, whereas PixelGen introduces perceptual losses to provide more effective guidance. (c) In the no-CFG setting, PixelGen surpasses latent diffusion baselines on ImageNet using only 80 epochs.}
    \label{fig:intro}
\end{figure}

Diffusion models~\cite{ddpm, ddim, adm} have achieved remarkable success in high-fidelity image generation, offering exceptional quality and diversity.
Research in this field generally follows two main directions: {latent diffusion} and {pixel diffusion}.
Latent diffusion models~\cite{ldm, dit, sit, flux2024} split generation into two stages. As illustrated in \cref{fig:intro}(a), a VAE first compresses images into a latent space, and a diffusion model then performs denoising in that space.
The performance of latent diffusion is largely constrained by the VAE, where the reconstruction quality limits the upper bound of generation, and the learned latent distribution further affects the convergence of diffusion training~\cite{vavae, repa_e}.
These works have shown that VAEs introduce low-level artifacts and representational bottlenecks for latent diffusion models.

Pixel diffusion models avoid these limitations by modeling raw pixels directly.
This end-to-end pipeline removes the need for latent representations and eliminates VAE-induced artifacts.
However, it is difficult for the diffusion model to learn a high-dimensional and complex velocity field in pixel space.
Recently, JiT~\cite{li2025jit} has improved pixel diffusion by predicting the clean image, \emph{i.e.}, $x$-prediction, rather than velocity or noise. Since the prediction target lies directly on the image manifold, $x$-prediction substantially simplifies optimization and improves generation quality.

\begin{figure}[t]
    \centering
    \includegraphics[width=0.95\linewidth]{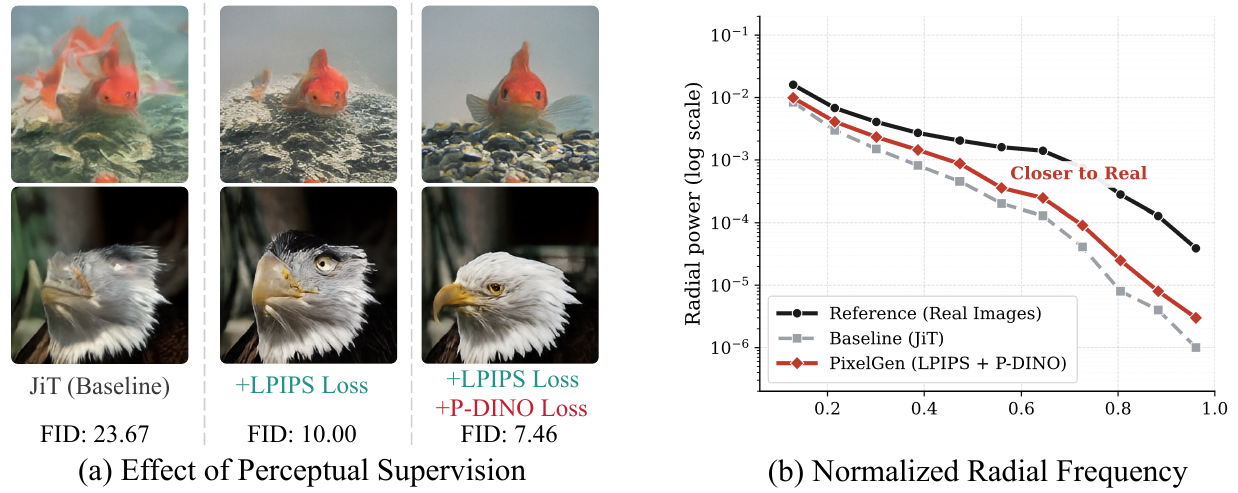}
\caption{(a)~LPIPS sharpens local textures over blurry samples of JiT, and P-DINO further improves global semantics. (b)~PixelGen better matches the real-image radial spectrum on $50{,}000$ samples.}
    \label{fig:empirical_ana}
\end{figure}

Despite this progress, a notable performance gap persists between JiT and latent diffusion. Beyond the prediction target, another key bottleneck lies in the training objective itself. Although $x$-prediction nominally constrains the output to lie on the image manifold, pixel diffusion still struggles to fit it. The standard pixel-wise loss supervises every pixel uniformly, so model capacity is largely spent on perceptually irrelevant components rather than salient structures. As shown in \cref{fig:empirical_ana}, this produces blurry samples and a clear spectral gap to real images across all frequency bands.
{To enable more efficient optimization, pixel diffusion needs perceptual supervision that emphasizes perceptually significant components instead of learning all pixels uniformly.}

Based on this insight, we propose PixelGen, a simple and effective training framework that augments $x$-prediction with perceptual supervision, as illustrated in \cref{fig:intro}~(b).
Since $x$-prediction directly estimates the clean image, it offers a natural interface to perceptual encoders such as LPIPS~\cite{lpips} and DINO~\cite{dinov2}, which are pretrained on clean images and only operate meaningfully in image space.
PixelGen introduces three coupled designs. First, an LPIPS-based perceptual loss~\cite{lpips} is applied on the predicted image to enhance local textures and fine details.
Second, a DINO-based perceptual loss, termed P-DINO, aligns last-layer features of a DINOv2 encoder to provide global semantic guidance.
Third, at high-noise timesteps, the predicted image is still blurry and lacks fine details, so forcing perceptual alignment with the clean image can over-constrain early denoising and reduce sample coverage, as reflected by lower recall~\cite{pr_recall} and further analyzed in \cref{sec:noise_gating}. We thus introduce a simple noise-gating strategy that enables the perceptual losses only at lower-noise timesteps, where the prediction is closer to a clean image.
PixelGen requires no latent representations, no VAEs, and no auxiliary stages.

We evaluate PixelGen on both class-to-image and text-to-image generation.
PixelGen achieves a leading FID score of 5.11 on ImageNet 256 without classifier-free guidance (CFG) using only 80 epochs, with only a small training overhead. In this no-CFG setting, it surpasses 
the strong latent diffusion model REPA~\cite{repa}, which achieves an FID of 5.90 with 800 training epochs. 
With CFG, PixelGen remains competitive, clearly improving over prior pixel diffusion models.
Beyond better metrics, the radial power spectrum in \cref{fig:empirical_ana}(b) shows that perceptual supervision pulls the generated distribution closer to real images, although a residual gap remains.
For text-to-image generation, PixelGen is efficiently pretrained from scratch in \textbf{only 6 days on 8$\times$H800 GPUs} and reaches a GenEval score of 0.79.

In summary, our contributions are as follows.
i) We propose PixelGen, a simple end-to-end pixel diffusion framework that augments $x$-prediction with perceptual supervision.
ii) We introduce three complementary designs for perceptual supervision: a local LPIPS loss, a global P-DINO loss, and a noise-gating strategy that disables high-noise supervision.
iii) PixelGen substantially narrows the gap between pixel and latent diffusion on ImageNet, and reaches GenEval 0.79 on text-to-image generation with only 6 days of training on 8$\times$H800 GPUs.

\section{Related Work}
This work is closely related to latent diffusion, pixel diffusion, and perceptual supervision. 

\noindent\textbf{Latent Diffusion.}
Latent diffusion trains diffusion models in a compact latent space learned by a VAE~\cite{ldm}. Compared with raw pixel space, this latent space significantly reduces spatial dimensionality, easing optimization and lowering computational cost~\cite{ldm, dcae}. Consequently, VAEs have become a fundamental component of diffusion models~\cite{dit, edm2, dod, flowdcn, dim, dmm, mdt, yao2024fasterdit, maskdit}. However, VAE training often involves adversarial objectives, which complicates the overall pipeline~\cite{wang2025pixnerd}. Poorly trained VAEs can also introduce decoding artifacts~\cite{sid, pixelflow} and representational bottlenecks.
Architecturally, DiT~\cite{dit} replaced the U-Net~\cite{uvit, adm} used in earlier latent diffusion, and SiT~\cite{sit} further validated DiT-style backbones with linear flow diffusion. Later works strengthen latent diffusion through representation alignment and joint optimization: REPA~\cite{repa} and REG~\cite{reg2025wu} align intermediate features to pretrained DINOv2~\cite{dinov2}, which is compatible with our framework and used in both our baseline and PixelGen. REPA-E~\cite{repa_e} jointly optimizes the VAE and DiT, but a changing latent space may be unstable during the diffusion training, whereas pixel diffusion keeps denoising targets fixed in pixel space. In other directions, VAVAE~\cite{vavae} and RAE~\cite{rae2025} improve autoencoders for faster training. DDT~\cite{decoupled_dit} proposes decoupled diffusion transformers. MeanFlow~\cite{geng2026mean} and Improved MeanFlow~\cite{imeanflow} pursue one-step generation with average velocity fields.

\noindent\textbf{Pixel Diffusion.}
Pixel diffusion avoids latent bottlenecks but has advanced more slowly because pixel space is high dimensional~\cite{adm, vdm, rdm, maskgit, simple_diffusion}. Early methods used multi-resolution diffusion. Relay Diffusion~\cite{rdm} trains separate scale-specific models, while PixelFlow~\cite{pixelflow} shares one model across scales but needs a complex denoising schedule. Recent work explores alternative designs, including fractal generation~\cite{fractal}, transformer-based normalizing flows~\cite{tarflow, zheng2025farmer}, neural-field velocity prediction~\cite{wang2025pixnerd}, self-supervised pretraining~\cite{lei2025epg}, and pixel decoders for high-frequency details~\cite{ma2025deco, chen2025dip, yu2025pixeldit}. JiT~\cite{li2025jit} predicts clean images instead of velocity to learn the low-dimensional image manifold. BiFlow~\cite{lu2025bidirectionalnormalizingflowdata} learns bidirectional data-noise mappings. pMF~\cite{pixelmeanflows} targets one-step pixel diffusion through pixel-space mean flows.

\noindent\textbf{Perceptual Supervision.}
Perceptual supervision replaces pixel-wise losses with feature-space objectives, emphasizing perceptually meaningful structure over exact RGB matches. It is widely used in autoencoders and GANs to reduce blur and sharpen details. LPIPS~\cite{lpips} is a common choice for improving local textures, while recent self-supervised encoders such as DINOv2~\cite{dinov2} can provide semantic features for global structure. Adversarial losses~\cite{styleganxl} can improve realism but are unstable and difficult to optimize for pixel diffusion, so we do not use them. Note that latent diffusion's VAE is also trained with LPIPS and GAN losses~\cite{ldm}.
In traditional latent diffusion, some early works have explored perceptual losses. Self-Perceptual loss~\cite{lin2025diffusionmodelperceptualloss} uses the diffusion model itself as the perceptual encoder, LPL~\cite{berrada2025boostingperceptual} uses VAE's decoder features and applies the loss at lower-noise timesteps, and Diffusion2GAN~\cite{kang2024diffusion2gan} combines perceptual and adversarial losses to distill diffusion models into conditional GANs. PixelGen shows that local LPIPS and global P-DINO supervision are complementary. We also find that, unlike LPL's stable recall across timesteps in latent diffusion, perceptual loss at high-noise timesteps hurts recall in pixel diffusion. It is because predictions are far from clean targets and can be over-constrained by feature matching.

\section{Methodology}
\label{sec:method}
In this section, we introduce PixelGen, a simple and effective training framework that augments $x$-prediction with perceptual supervision.
We first present an overview of PixelGen in \cref{sec:overview}. We then introduce two complementary perceptual losses: an LPIPS loss for local textures in \cref{sec:LPIPS} and a P-DINO loss for global semantics in \cref{sec:DINO}. Finally, we describe a simple noise-gating strategy in \cref{sec:noise_gating}.

\subsection{Overview}
\label{sec:overview}

In pixel diffusion, the model output can be parameterized as noise ($\epsilon$), velocity ($v$), or image ($x$), corresponding to $\epsilon$-, $v$-, and $x$-prediction, respectively. JiT~\cite{li2025jit} recently simplified the target by replacing the widely used $v$-prediction with $x$-prediction, substantially improving pixel diffusion.

However, a clear gap remains between JiT and strong latent diffusion models. We attribute this gap in part to the inefficiency of uniform pixel-wise supervision, which spends capacity on perceptually insignificant signals such as sensor noise and imperceptible details~\cite{ldm} while giving limited emphasis to structures that determine visual quality. As a result, generated samples can still drift from the real image manifold, as shown in \cref{fig:empirical_ana}. Our key insight is that pixel diffusion should not learn all pixels equally. Instead, it should receive perceptual supervision that prioritizes visually meaningful components and pulls predictions toward the real image manifold. Since $x$-prediction directly estimates the clean image, it provides a natural interface for perceptual encoders pretrained on clean images. We next introduce image prediction with flow matching, and then present the proposed perceptual supervision.

\noindent\textbf{Image Prediction.} Following JiT~\cite{li2025jit}, we adopt image prediction, \emph{i.e.}, $x$-prediction, to provide a stable target across noise levels. Given a noisy image $x_t$ at time $t \in [0,1]$, the diffusion transformer $\text{net}_{\theta}$ predicts the clean image $x_\theta$ as:
\begin{equation}
    x_\theta = \operatorname{net}_{\theta}(x_t, t, c),
\end{equation}
where $c$ denotes conditional information, such as class labels or text embeddings. The noisy input $x_t$ is constructed by linearly interpolating between the ground-truth image $x$ and Gaussian noise $\epsilon \sim \mathcal{N}(0, I)$:
\begin{equation}
    x_t = t x + (1 - t)\epsilon.
\end{equation}

\noindent\textbf{Velocity Conversion.}
To retain the sampling advantages of flow matching, we convert the predicted image $x_\theta$ into a velocity $v_\theta$ following JiT~\cite{li2025jit}:
\begin{equation}
    v_\theta = \frac{x_\theta - x_t}{1 - t},
\end{equation}
while the ground-truth velocity $v$ can be represented as:
\begin{equation}
    v = \frac{x - x_t}{1 - t} = x - \epsilon.
\end{equation}
The resulting flow matching objective is:
\begin{equation}
\label{eq:fm_loss}
    \mathcal{L}_{\text{FM}} =
    \mathbb{E}_{t,x,\epsilon}
    \left\| v_\theta - v \right\|^2
    =
    \mathbb{E}_{t,x,\epsilon}
    \left\| \frac{x_\theta - x}{1 - t} \right\|^2.
\end{equation}
This formulation combines the target of $x$-prediction with the sampling advantages of flow matching.

\noindent\textbf{Perceptual Supervision.}
Although $x$-prediction simplifies the training objective, the diffusion loss still supervises pixels uniformly and is therefore misaligned with perceptual quality. PixelGen addresses this by applying two complementary perceptual losses directly to the predicted clean image $x_\theta$: LPIPS for local textures and fine details, and P-DINO for global semantics.
This image-space supervision is most useful when $x_\theta$ is sufficiently close to a clean image, because clean-image encoders can over-constrain high-noise predictions. We therefore introduce a simple gate $g(t)$ that enables perceptual losses only in the low-noise regime.
Together with the widely used REPA loss~\cite{repa}, which encourages alignment of intermediate representations with a pretrained DINOv2 encoder, the final training objective is:
\begin{equation}
    \mathcal{L}
    =
    \mathcal{L}_{\text{FM}}
    +
    \lambda_1 g(t)\mathcal{L}_{\text{LPIPS}}
    +
    \lambda_2 g(t)\mathcal{L}_{\text{P-DINO}}
    + \mathcal{L}_{\text{REPA}}
    ,
\end{equation}
where $\lambda_1$ and $\lambda_2$ balance the diffusion objective and perceptual supervision. The gate $g(t)$ disables perceptual losses at high-noise timesteps, as detailed in \cref{sec:noise_gating}. This end-to-end training enables PixelGen to better fit the real image manifold without VAEs.

\begin{figure}[t]
    \centering
    \includegraphics[width=0.95\linewidth]{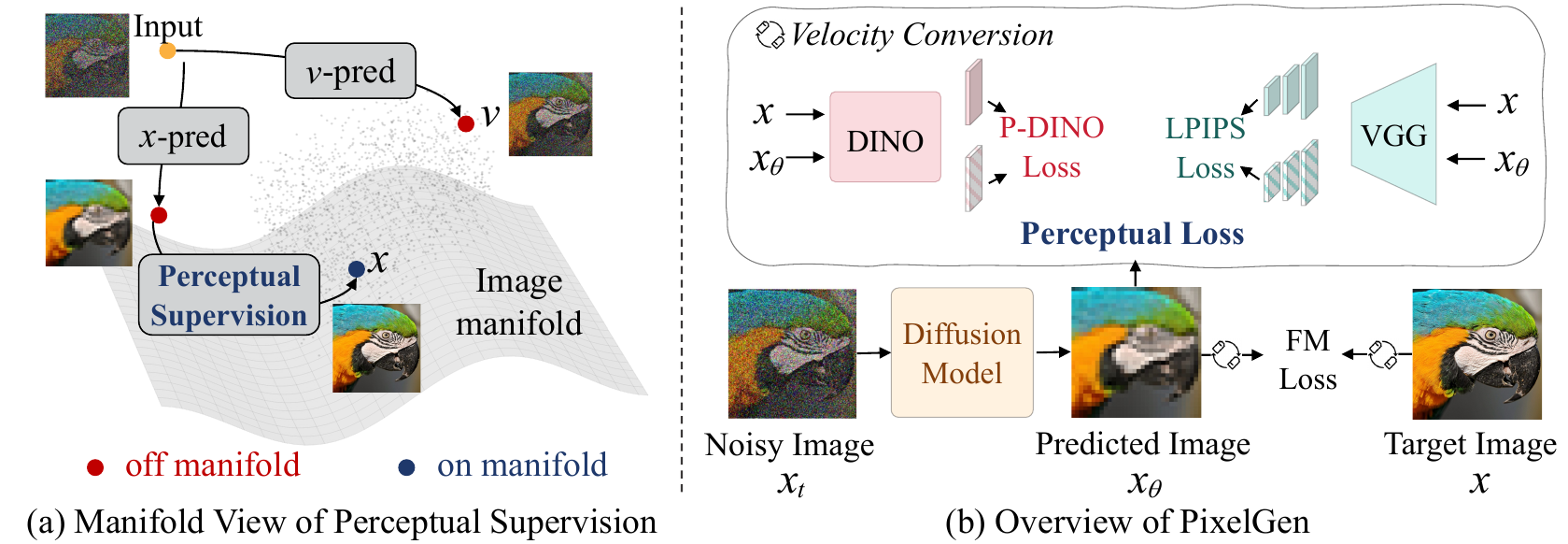}
    \caption{Overview of PixelGen. (a)~$v$-prediction and pixel-wise $x$-prediction struggle to fit the image manifold, while perceptual supervision guides $x$-predictions toward it. (b)~PixelGen combines flow matching with LPIPS for local texture and P-DINO for global semantics.}
    \label{fig:architecture}
\end{figure}

\subsection{LPIPS Loss}
\label{sec:LPIPS}

High-quality image generation requires sharp details and realistic local textures, which are not well captured by pixel-wise losses. To address this, we incorporate the Learned Perceptual Image Patch Similarity (LPIPS) loss~\cite{lpips}.
LPIPS measures perceptual similarity by comparing multi-level feature activations extracted from a frozen pretrained VGG network $f_{\text{VGG}}$. The LPIPS loss can be written as:
\begin{equation}
    \mathcal{L}_{\text{LPIPS}}
    =
    \sum_l
    \left\|
    w_l \odot
    \left(
    f_{\text{VGG}}^l(x_\theta)
    -
    f_{\text{VGG}}^l(x)
    \right)
    \right\|_2^2,
\end{equation}
where $l$ indexes VGG layers, and $w_l$ denotes the learned per-channel weighting vector for layer $l$. For simplicity, we omit spatial averaging over feature maps and channel-wise normalization in the formulation.

By minimizing $\mathcal{L}_{\text{LPIPS}}$, PixelGen learns perceptually important local patterns instead of matching exact pixel values. As shown in \cref{fig:empirical_ana}(a), adding LPIPS to the JiT baseline sharpens local textures and fine details. Quantitatively, it reduces FID from 23.67 to 10.00 on ImageNet without classifier-free guidance, indicating that local perceptual supervision is much more effective than uniform pixel-wise matching for correcting over-smoothed predictions.

\subsection{P-DINO Loss}
\label{sec:DINO}
While LPIPS provides strong local perceptual supervision, local patterns alone are insufficient for high-fidelity generation. 
We therefore introduce a Perceptual DINO (P-DINO) loss to provide global semantic guidance.

Specifically, we extract patch-level features using a frozen DINOv2-B~\cite{dinov2} encoder $f_{\text{DINO}}$.
Let $f_{\text{DINO}}^p(\cdot)$ denote the feature of patch $p$.
We align the predicted image ${x}_\theta$ and the ground-truth image ${x}$ using cosine similarity:
\begin{equation}
    \mathcal{L}_{\text{P-DINO}}
    =
    \frac{1}{|\mathcal{P}|}
    \sum_{p\in\mathcal{P}}
    \left(
    1
    -
    \cos\bigl(
    f_{\text{DINO}}^p(x_\theta),
    f_{\text{DINO}}^p(x)
    \bigr)
    \right),
\end{equation}
where $\mathcal{P}$ denotes the set of all patches.
The P-DINO loss provides global semantic guidance by aligning high-level representations, encouraging the predicted image to be consistent with the overall scene layout and object semantics.
Together with the local LPIPS loss, it enables PixelGen to balance global semantics and local realism, pulling the predicted images closer to the real image manifold.

P-DINO complements LPIPS by improving the global semantics that local texture supervision alone cannot fully capture. As shown in \cref{fig:empirical_ana}(a), adding P-DINO on top of LPIPS yields clearer object semantics and more coherent global structures, further reducing FID from 10.00 to 7.46. With these two complementary perceptual losses, as illustrated in \cref{fig:empirical_ana}(b), spectral alignment is consistently improved over the JiT baseline, suggesting that perceptual supervision pulls pixel diffusion closer to the real image manifold.

\subsection{Noise Gating}
\label{sec:noise_gating}

Although LPIPS and P-DINO provide useful perceptual supervision, applying them uniformly across all timesteps can be harmful. At high-noise timesteps, the predicted image $x_\theta$ is still blurry and lacks fine details, so clean-image feature matching can over-constrain early denoising. We provide an empirical analysis of perceptual gradients across timesteps in Appendix~\ref{appendix:noise_gating}, which suggests that high-noise perceptual gradients are comparatively large relative to the flow matching gradient.

We therefore use a simple timestep gate for perceptual supervision:
\begin{equation}
    g(t)
    =
    \mathbf{1}\left[t \ge \tau\right],
\end{equation}
where $t=0$ corresponds to pure noise and $t=1$ corresponds to clean data under the interpolation in \cref{sec:overview}. We set $\tau=0.3$, disabling perceptual losses during the first 30\% high-noise timesteps and activating them during the later 70\% low-noise timesteps. This strategy does not change the diffusion objective or the sampling process. It only limits perceptual supervision when $x_\theta$ is too noisy.

The ablation in \cref{exp-ablation-T-clamp} shows that applying perceptual losses at high-noise timesteps hurts recall~\cite{pr_recall}, which measures the coverage of real images by generated samples. Noise gating improves recall with only a small trade-off in FID and precision, suggesting that perceptual losses are most useful after the prediction enters a reasonably clean regime. Together with the LPIPS and P-DINO analyses above, this supports the central design of PixelGen, which applies perceptual supervision at low-noise timesteps and relies on the flow matching objective elsewhere.

\section{Experiments}
We evaluate PixelGen on ImageNet $256\times256$ and GenEval~\cite{geneval}. \cref{sec-exp-baseline} compares PixelGen with latent and pixel diffusion baselines under a 200K-step no-CFG setting on ImageNet, and \cref{sec-exp-ablation} ablates the main components. \cref{sec-exp-class2img} reports class-to-image results on ImageNet with FID~\cite{fid}, Inception Score, precision, and recall~\cite{pr_recall}. \cref{sec-exp-text2img} reports text-to-image results on GenEval.

\subsection{Comparison with Baselines}
\label{sec-exp-baseline}
\noindent\textbf{Setup.}
We first compare PixelGen with latent and pixel diffusion baselines under the same training setting. All models are trained on ImageNet $256\times256$ for \textit{200K training steps} with a DiT-L backbone~\cite{dit}. Following prior work~\cite{dit,wang2025pixnerd,ma2025deco}, we use a global batch size of 256, AdamW with a constant learning rate of $1\times10^{-4}$, and log-normal timestep sampling. For fair comparison, we apply REPA loss~\cite{repa} to all models except DiT-L/2 and PixelFlow-L/4~\cite{pixelflow}. The DiT patch size is 16, and JiT~\cite{li2025jit} with REPA loss serves as our baseline.
For inference, we use 50 Euler steps without CFG. All experiments are run on one node with 8$\times$H800 GPUs.

\begin{figure*}[t]
    \centering
    \includegraphics[width=1\linewidth]{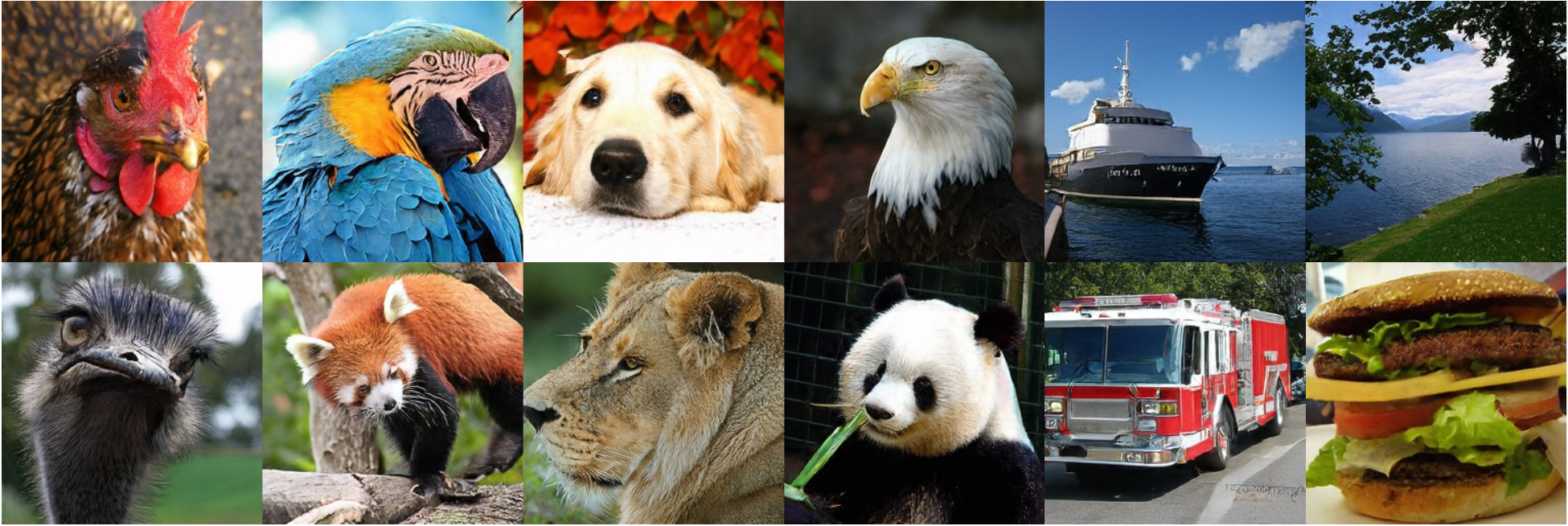}
    \caption{Class-to-image samples from PixelGen with CFG on ImageNet $256\times256$.}
    \label{fig-c2i-visualization}
\end{figure*}

\begin{table}[t]
    \centering
    \caption{ImageNet 256 results after \emph{200K} training steps without CFG. The inference adopts 50 Euler steps. REPA loss is used for all models except DiT-L/2 and PixelFlow-L. Latent diffusion models require an extra VAE with 86M parameters. $\dagger$ denotes online VAE encoding.}
    \label{tab-baseline-comparison}
    \footnotesize
    \setlength{\tabcolsep}{2.5pt}
    \resizebox{\linewidth}{!}{%
    \begin{tabular}{c|l|c|ccc|c|cccc}
    \toprule
     & & & \multicolumn{3}{c|}{Training} & Inference & \multicolumn{4}{c}{Generation Metrics} \\
     \multicolumn{1}{c|}{} & Method & Params & Speed (s/it) & Hours & Mem (GB)& (s/image) & FID$\downarrow$ & IS$\uparrow$ & Prec.$\uparrow$ & Rec.$\uparrow$ \\ \midrule
    \multirow{3}{*}{\rotatebox[origin=c]{90}{{Latent}}}
     & DiT-L/2 $\dagger$~\cite{dit}  & 458+86M & 0.34 & 18.9 & 28.5 & 0.43 & 41.93 & 36.5  & 0.52 & 0.59 \\
     & REPA-L/2 ~\cite{repa}& 458+86M & 0.17    &  9.7   & 22.0    & 0.43    & 16.14 & 87.3  & 0.65 & 0.63 \\
     & DDT-L/2  $\dagger$~\cite{decoupled_dit}  & 458+86M & 0.28    & 15.5    & 30.1    &  0.43   & 10.00 & 112.9 & 0.67 & \textbf{0.65} \\
    \hline
    \multirow{7}{*}{\rotatebox[origin=c]{90}{{Pixel}}}
     & PixNerd-L/16~\cite{wang2025pixnerd}   & 459M & 0.20  & 11.0  & 29.2 & 0.33            & 37.49 & 43.0 & 0.46 & 0.62 \\
     & PixelFlow-L/4~\cite{pixelflow}  & 459M & 1.27  & 70.8  & 73.2 & 4.60 & 54.33 & 24.7 & 0.43 & 0.58 \\
     & DeCo-L/16~\cite{ma2025deco}      & 426M & 0.19  & 10.6  & 27.5 & 0.32            & 31.35 & 48.4 & 0.51 & \textbf{0.65} \\
     & JiT-L/16 {(Baseline)} ~\cite{li2025jit} & 459M & 0.19 & 10.6 & 29.5 & 0.32 & 23.67 & 64.1 & 0.55 & 0.63 \\
    \rowcolor{blue!10} \cellcolor{white} & PixelGen-L/16 (\textit{100K}) & 459M & 0.20 & 5.5  & 33.9 & 0.32 & 10.50         & 106.2              & 0.70             & 0.58    \\
    \rowcolor{blue!10} \cellcolor{white} & PixelGen-L/16        & 459M & 0.20 & 11.1 & 33.9 & 0.32 & \textbf{7.53} & \textbf{131.7} & \textbf{0.72} & 0.60 \\
    \bottomrule
    \end{tabular}%
    }
\end{table}

\noindent\textbf{Detailed Comparisons.} 
\cref{tab-baseline-comparison} shows that PixelGen outperforms both latent and pixel-based diffusion baselines under the same \textit{200K} training setting. 
Compared with the JiT baseline~\cite{li2025jit}, PixelGen reduces FID from 23.67 to 7.53. This result suggests that perceptual losses provide more effective supervision than the standard pixel diffusion loss, guiding predicted images closer to the real image manifold. PixelGen also improves over recent pixel diffusion methods such as PixNerd~\cite{wang2025pixnerd} and DeCo~\cite{ma2025deco} under the same protocol.
Under the 200K-step no-CFG setting, PixelGen is also competitive with strong latent diffusion baselines, achieving an FID of 7.53 compared with 10.00 for DDT-L/2~\cite{decoupled_dit} and 16.14 for REPA-L/2~\cite{repa}. Since latent diffusion can also benefit from perceptual supervision~\cite{berrada2025boostingperceptual,lin2025diffusionmodelperceptualloss}, we do not claim that pixel diffusion is intrinsically superior. Our goal is instead to show that perceptual losses can be effectively integrated into a pixel diffusion framework, improving one-stage end-to-end generation. The comparison is fair, since latent diffusion's VAE is also trained with LPIPS and GAN perceptual losses~\cite{ldm}.

These gains come with small overhead. As shown in \cref{tab-baseline-comparison}, perceptual supervision increases per-step time from 0.19s to 0.20s and memory from 29.5\,GB to 33.9\,GB over JiT, while keeping the same inference cost of 0.32\,s/image. PixelGen reaches 10.50 FID in 5.5 hours and further improves to 7.53 FID at 200K steps. Compared with latent diffusion, PixelGen removes VAE training and storage while achieving lower FID and faster inference than DiT-L/2, with 0.32\,s/image versus 0.43\,s/image.

\subsection{Class-to-Image Generation}
\label{sec-exp-class2img}
\noindent\textbf{Setup.}
For class-to-image generation on ImageNet, we train PixelGen-XL with 676M parameters at $256\times256$ resolution for 160 epochs. During inference, we follow JiT and use the Heun sampler~\cite{heun1900neue} with 50 steps. When CFG is enabled, we use guidance interval~\cite{interval_guidance}.

\noindent\textbf{Results without CFG.}
\cref{tab-c2i-comparison-wo-cfg} reports class-to-image performance without CFG. This setting directly tests whether the model captures the underlying image distribution.
PixelGen-XL/16 achieves an \emph{FID of 5.11 using only 80 training epochs}, improving over strong latent diffusion baselines such as REPA-XL/2 with FID 5.90 at 800 epochs and DDT-XL/2 with FID 6.27 at 400 epochs. It also improves over prior pixel diffusion models. For example, DeCo-XL/16 reaches FID 14.88 with 320 epochs, while PixelGen reduces FID by more than 60\% using one quarter as many epochs.
PixelGen also obtains competitive Inception Score, precision, and recall. These results suggest that, in our no-CFG setting, perceptual supervision helps end-to-end pixel diffusion fit the real image manifold without a separately trained VAE.

\noindent\textbf{Results with CFG.}
\cref{tab-c2i-comparison-with-cfg} reports results with CFG. With 160 training epochs and 50 Heun inference steps, PixelGen achieves an FID of 1.83. It improves over recent pixel diffusion models such as DeCo-XL/16~\cite{ma2025deco} and JiT-H/16~\cite{li2025jit}, which use 320 and 600 training epochs.
A small gap remains to the leading latent baseline REPA-XL/2, which reaches FID 1.42 at 800 epochs. We attribute this gap partly to the interaction between CFG and perceptual losses in sample coverage. CFG improves precision while reducing recall by moving samples away from low-density regions~\cite{cfg, interval_guidance}, and our perceptual losses push coverage in the same direction as shown in \cref{exp-ablation-T-clamp}. We leave improved pixel-space samplers and CFG strategies tailored to perceptual supervision for future work.

\begin{table}[t]
    \centering
    \begin{minipage}[t]{0.48\linewidth}
    \centering
    \small
    \captionof{table}{Class-to-image generation performance without CFG on ImageNet 256. PixelGen follows JiT~\cite{li2025jit} and uses 50 Heun~\cite{heun1900neue} steps, while latent diffusion models use 250 Euler steps.}
    \label{tab-c2i-comparison-wo-cfg}
    \setlength{\tabcolsep}{4pt}
    \renewcommand{\arraystretch}{1.1}
    \resizebox{\linewidth}{!}{%
    \begin{tabular}{c|l|c|cccc}
    \toprule
    \multicolumn{2}{c|}{\textbf{Method}} & \textbf{Epochs} & \textbf{FID}$\downarrow$ & \textbf{IS}$\uparrow$ & \textbf{Prec.}$\uparrow$ & \textbf{Rec.}$\uparrow$ \\
    \midrule
    \multirow{10}{*}{\rotatebox[origin=c]{90}{{Latent}}} 
     & LDM~\cite{ldm} & 170 & 10.56 & 103.5 & 0.71 & 0.62 \\
     & DiT-XL/2~\cite{dit} & 1400 & 9.62 & 121.5 & 0.67 & 0.67 \\
     & SiT-XL/2~\cite{sit} & 1400 & 8.61 & 131.7 & 0.68 & 0.67 \\
     & FlowDCN~\cite{flowdcn} & 400 & 8.36 & 122.5 & 0.69 & 0.65 \\
     & FasterDiT~\cite{yao2024fasterdit} & 400 & 7.91 & 131.3 & 0.67 & 0.69 \\
     & DDT-XL/2~\cite{decoupled_dit} & 80 & 6.62 & 135.2 & 0.69 & 0.67 \\
     & DDT-XL/2~\cite{decoupled_dit} & 400 & 6.27 & 154.7 & 0.69 & 0.67 \\
     & MDT~\cite{mdt} & 1300 & 6.23 & 143.0 & 0.71 & 0.65 \\
     & REPA-XL/2~\cite{repa} & 800 & 5.90 & 157.8 & 0.70 & \textbf{0.69} \\
     & MaskDiT~\cite{maskdit} & 1600 & 5.69 & \textbf{177.9} & \textbf{0.74} & 0.60 \\
    \midrule
    \multirow{5}{*}{\rotatebox[origin=c]{90}{{Pixel}}}
     & ADM~\cite{adm} & 400 & 10.94 & - & 0.69 & 0.63 \\
     & PixelFlow-XL~\cite{pixelflow} & 320 & 12.23 & 103.3 & 0.63 & 0.66 \\
     & PixNerd-XL~\cite{wang2025pixnerd} & 320 & 15.61 & 88.9 & 0.59 & 0.68 \\
     & {DeCo-XL/16}~\cite{ma2025deco} & 320 & 14.88 & 88.2 & 0.60 & 0.68 \\
    \rowcolor{blue!10} \cellcolor{white} & {PixelGen-XL/16} & 80 & \textbf{5.11} & {159.2} & 0.72 & 0.63 \\
    \bottomrule
    \end{tabular}%
    }
    \end{minipage}%
    \hfill
    \begin{minipage}[t]{0.48\linewidth}
    \centering
    \footnotesize
    \captionof{table}{Class-to-image generation performance {with CFG} on ImageNet 256.
    }
    \label{tab-c2i-comparison-with-cfg}
    \setlength{\tabcolsep}{4pt}
    \renewcommand{\arraystretch}{1.1}
    \resizebox{\linewidth}{!}{%
    \begin{tabular}{c|l|c|cccc}
    \toprule
    \multicolumn{2}{c|}{\textbf{Method}} & \textbf{Epochs} & \textbf{FID}$\downarrow$ & \textbf{IS}$\uparrow$ & \textbf{Prec.}$\uparrow$ & \textbf{Rec.}$\uparrow$ \\
    \midrule
    \multirow{6}{*}{\rotatebox[origin=c]{90}{{Latent}}} 
     & MaskDiT~\cite{maskdit} & 1600 & 2.28 & 276.6 & 0.80 & 0.61 \\
     & DiT-XL/2~\cite{dit} & 1400 & 2.27 & 278.2 & 0.83 & 0.57 \\
     & SiT-XL/2~\cite{sit} & 1400 & 2.06 & 284.0 & 0.83 & 0.59 \\
      & MDT~\cite{mdt} & 1300 & 1.79 & 283.0 & 0.81 & 0.61 \\
     & REPA-XL/2~\cite{repa} & 200 & {1.96} & 264.0 & 0.82 & {0.60} \\
     & REPA-XL/2~\cite{repa} & 800 & \textbf{1.42} & 305.7 & 0.80 & \textbf{0.64} \\
    \midrule
    \multirow{10}{*}{\rotatebox[origin=c]{90}{{Pixel}}}
    & ADM~\cite{adm} & 400 & 4.59 & 186.7 & \textbf{0.82} & 0.52 \\
     & SimpleDiffusion~\cite{simple_diffusion} & 800 & 2.44 & 256.3 & - & - \\
     & RDM~\cite{rdm} & 400 & 1.99 & 260.4 & 0.81 & 0.58 \\
     & FractalMAR-H & 600 & 6.15 & \textbf{348.9} & 0.81 & 0.46 \\
     & EPG-XL~\cite{lei2025epg} & 800 & 2.04 & 283.2 & 0.80 & 0.61 \\
     & PixelFlow-XL~\cite{pixelflow} & 320 & 1.98 & 282.1 & 0.81 & 0.60 \\
     & {DiP-XL/16}~\cite{chen2025dip} & 320 & {1.98} & 282.9 & 0.80 & {0.62} \\
     & PixNerd-XL~\cite{wang2025pixnerd} & 320 & 1.95 & 298.0 & 0.80 & 0.60 \\
     & {DeCo-XL/16}~\cite{ma2025deco} & 320 & {1.90} & 303.0 & 0.80 & {0.61} \\
     & {JiT-H/16}~\cite{li2025jit} & 600 & 1.86 & 303.4 & 0.78 & 0.62 \\
    \rowcolor{blue!10} \cellcolor{white} & {PixelGen-XL/16} & 160 & 1.83 & 293.6 & 0.79 & 0.63 \\
    \bottomrule
    \end{tabular}%
    }
    \end{minipage}
\end{table}

\noindent\textbf{Qualitative Results.}
\cref{fig-c2i-visualization} shows samples generated by PixelGen-XL/16 on ImageNet $256\times256$ with CFG, which exhibit accurate class semantics, sharp textures, and coherent global structures across animals, objects, and natural scenes.

\begin{table*}[t]
\centering
\caption{Text-to-image generation on GenEval~\cite{geneval} at a 512$\times$512 resolution.}
\label{tab-my-table}
\footnotesize
\setlength{\tabcolsep}{4pt}
\begin{tabular}{l|c|ccccccc}
\toprule
 & Diffusion & \multicolumn{7}{c}{GenEval} \\   
Method     & Params        & Sin.Obj. & Two.Obj & Counting & Colors & Pos & Color.Attr. & Overall$\uparrow$ \\ \midrule
PixArt-$\alpha$~\cite{chen2023pixartalpha} & 0.6B & 0.98 & 0.50  & 0.44  & 0.80 &  0.08 &  0.07 & 0.48 \\
SD3~\cite{sd3}  & 8B        & 0.98                & 0.84             & 0.66           & 0.74         & 0.40           & 0.43                    & 0.68         \\
FLUX.1-dev~\cite{flux2024} &  12B  & 0.99                & 0.81             & {0.79}           & 0.74         & 0.20           & 0.47                    & 0.67          \\
DALL-E 3~\cite{Betker2023ImprovingIG} & - & 0.96 & 0.87 &  0.47   &   0.83 &  0.43&  0.45  & 0.67 \\
OmniGen2~\cite{wu2025omnigen2}      &   4B  & {1}                   & {0.95}             & 0.64           & 0.88         & 0.55           & 0.76                    & 0.80      \\ 
\hline
PixelFlow-XL/4~\cite{pixelflow} &  882M &      -        & -                & -              & -            & -              & -                       & 0.60      \\ 
PixNerd-XXL/16~\cite{wang2025pixnerd} &  1.2B &      0.97        & 0.86              & 0.44         & 0.83            & 0.71             & 0.53                      & 0.73      \\

\rowcolor{blue!10} {PixelGen-XXL/16} & 1.1B & 0.99 & 0.88  & 0.59  & {0.90} & {0.70}  &  {0.70}   &  {0.79}   \\
        
        \bottomrule
\end{tabular}%

\end{table*}

\subsection{Text-to-Image Generation}
\label{sec-exp-text2img}

\noindent\textbf{Setup.}
For text-to-image generation, we train on about 36M pretraining images and 60k high-quality instruction-tuning samples from BLIP3o~\cite{blip3o}. We use Qwen3-1.7B~\cite{yang2025qwen3} as the text encoder. Following Fluid~\cite{fluid}, we train several transformer layers on top of the frozen text features to improve feature alignment~\cite{fluid}. The total batch size is 1536 for $256\times256$ pretraining and 512 for $512\times512$ pretraining. Following previous work~\cite{wang2025pixnerd}, we pretrain PixelGen-XXL at $256\times256$ for 200K steps and then at $512\times512$ for 80K steps. We further fine-tune it on BLIP3o-60k for 40K steps at $512\times512$. We use gradient clipping for stable training, adopt the Adams-2nd solver with 25 steps as the default sampler, and set the CFG scale to 4.0. \emph{The full training takes about 6 days on 8$\times$H800 GPUs.} 

\noindent\textbf{Main Results.}
We evaluate PixelGen on text-to-image generation to test its scalability and generalization. Quantitative results on GenEval are reported in \cref{tab-my-table}.
PixelGen-XXL achieves an overall GenEval score of 0.79, which is competitive with recent large-scale diffusion models such as FLUX.1-dev~\cite{flux2024} with 0.67 while using fewer parameters and less compute. We do not claim superiority over these systems, since training data, model size, and compute differ substantially. Instead, this result indicates that end-to-end pixel diffusion with perceptual supervision can reach a competitive GenEval score under a much smaller training cost.
PixelGen also improves over recent pixel diffusion methods such as PixNerd, suggesting that perceptual supervision is useful for end-to-end pixel diffusion beyond ImageNet class-to-image generation.

\noindent\textbf{Practical efficiency.}
PixelGen-XXL is pretrained from scratch in only \emph{6 days on 8$\times$H800 GPUs}, without an extra VAE, providing a simple framework for future end-to-end pixel diffusion research.

\renewcommand{\thesubtable}{{\alph{subtable}}}

\subsection{Ablation Experiments}
\label{sec-exp-ablation}
We study the key components of PixelGen on ImageNet $256\times256$ under the same setup as \cref{sec-exp-baseline}.
We first ablate each component and then analyze the main hyperparameters.

\begin{table*}[t]
\centering
\caption{Ablation experiments of PixelGen on ImageNet $256\times256$. \textcolor{blue!40}{Blue background} indicates the default configuration. All models are trained for 200K steps with the same setup as \cref{sec-exp-baseline}.}
\label{exp-tab-ablation}
\scriptsize
\setlength{\tabcolsep}{2pt}
\renewcommand{\arraystretch}{1.1}

\begin{minipage}[t]{0.40\linewidth}
\centering
\subfloat[Effectiveness of each component.\label{exp-ablation-component}]{%
\begin{tabular}{l|cccc}
\toprule
Method & FID$\downarrow$ & IS$\uparrow$ & Prec.$\uparrow$ & Rec.$\uparrow$ \\
\midrule
Baseline JiT & 23.67 & 64.13 & 0.55 & 0.63 \\
+ LPIPS Loss & 10.00 & 113.16 & 0.70 & 0.59 \\
\quad + P-DINO & 7.46 & 137.95 & 0.73 & 0.58 \\
\rowcolor{blue!10} \quad\quad + Noise-Gate & 7.53 & 131.70 & 0.72 & 0.60 \\
\bottomrule
\end{tabular}%
}
\end{minipage}%
\hfill
\begin{minipage}[t]{0.28\linewidth}
\centering
\subfloat[Weight $\lambda_1$ of LPIPS Loss.\label{exp-ablation-lpips-weight}]{%
\begin{tabular}{c|cccc}
\toprule
Weight & FID$\downarrow$ & IS$\uparrow$ & Prec.$\uparrow$ & Rec.$\uparrow$ \\
\midrule
0.05 & 10.89 & 106.95 & 0.68 & 0.61 \\
\rowcolor{blue!10} 0.1 & 10.00 & 113.16 & 0.70 & 0.59 \\
0.5 & 9.36 & 122.34 & 0.71 & 0.58 \\
1.0 & 10.12 & 117.75 & 0.71 & 0.57 \\
\bottomrule
\end{tabular}%
}
\end{minipage}%
\hfill
\begin{minipage}[t]{0.28\linewidth}
\centering
\subfloat[Weight $\lambda_2$ of P-DINO Loss.\label{exp-ablation-dino-weight}]{%
\begin{tabular}{c|cccc}
\toprule
Weight & FID$\downarrow$ & IS$\uparrow$ & Prec.$\uparrow$ & Rec.$\uparrow$ \\
\midrule
0.005 & 8.11 & 128.86 & 0.72 & 0.59 \\
\rowcolor{blue!10} 0.01 & 7.46 & 137.95 & 0.73 & 0.58 \\
0.02 & 6.84 & 149.23 & 0.74 & 0.57 \\
0.04 & 6.62 & 157.78 & 0.73 & 0.57 \\
\bottomrule
\end{tabular}%
}
\end{minipage}


\begin{minipage}[t]{0.40\linewidth}
\centering
\subfloat[Compatibility with REPA~\cite{repa}.\label{exp-ablation-repa}]{%
\begin{tabular}{l|c|cccc}
\toprule
Method & REPA & FID$\downarrow$ & IS$\uparrow$ & Prec.$\uparrow$ & Rec.$\uparrow$ \\
\midrule
Baseline & \checkmark & 23.67 & 64.13 & 0.55 & 0.63 \\
\rowcolor{blue!10} PixelGen & \checkmark & 7.53 & 131.70 & 0.72 & 0.60 \\
Baseline & $\times$ & 34.85 & 42.45 & 0.50 & 0.61 \\
PixelGen & $\times$ & 11.81 & 105.30 & 0.67 & 0.60 \\
\bottomrule
\end{tabular}%
}
\end{minipage}%
\hfill
\begin{minipage}[t]{0.28\linewidth}
\centering
\subfloat[Selected DINO layer.\label{exp-ablation-depth}]{%
\begin{tabular}{c|cccc}
\toprule
Depth & FID$\downarrow$ & IS$\uparrow$ & Prec.$\uparrow$ & Rec.$\uparrow$ \\
\midrule
6 & 12.65 & 95.92 & 0.68 & 0.58 \\
9 & 10.01 & 111.51 & 0.71 & 0.58 \\
\rowcolor{blue!10} 12 & 7.46 & 137.95 & 0.73 & 0.58 \\
6,9,12 & 10.01 & 111.50 & 0.71 & 0.58 \\
\bottomrule
\end{tabular}%
}
\end{minipage}%
\hfill
\begin{minipage}[t]{0.28\linewidth}
\centering
\subfloat[Noise-Gating threshold.\label{exp-ablation-T-clamp}]{%
\begin{tabular}{c|cccc}
\toprule
Thres. & FID$\downarrow$ & IS$\uparrow$ & Prec.$\uparrow$ & Rec.$\uparrow$ \\
\midrule
0.0 & 7.46 & 137.95 & 0.73 & 0.58 \\
0.1 & 7.42 & 136.95 & 0.72 & 0.58 \\
\rowcolor{blue!10} 0.3 & 7.53 & 131.71 & 0.72 & 0.60 \\
0.6 & 10.72 & 109.50 & 0.69 & 0.60 \\
\bottomrule
\end{tabular}%
}
\end{minipage}
\end{table*}

\noindent\textbf{Effectiveness of each component.}
\cref{exp-ablation-component} summarizes the contribution of each component.
Starting from JiT, adding LPIPS substantially improves FID and IS, supporting the importance of local perceptual supervision in pixel space.
Adding P-DINO further improves global structure and brings additional gains.
However, applying perceptual losses at all timesteps reduces recall~\cite{pr_recall}, suggesting a fidelity-coverage trade-off when high-noise predictions are strongly aligned to clean-image features.
We therefore use noise gating to disable perceptual losses during the first 30\% high-noise timesteps. This improves recall with only a small trade-off in FID and precision.

\noindent\textbf{Loss weight of LPIPS.}
We vary the LPIPS weight $\lambda_1$ in \cref{exp-ablation-lpips-weight}.
A small weight of 0.05 gives weaker FID, while larger weights of 0.5 slightly reduce recall.
We set $\lambda_1=0.1$ as a balanced choice.

\noindent\textbf{Loss weight of P-DINO.}
We ablate the P-DINO weight $\lambda_2$ in \cref{exp-ablation-dino-weight}.
Increasing $\lambda_2$ generally improves FID, but larger weights can reduce recall.
We set $\lambda_2=0.01$ to balance FID and recall.

\noindent\textbf{Compatibility with REPA.}
\cref{exp-ablation-repa} studies compatibility with the REPA~\cite{repa} loss.
REPA improves JiT from FID 34.85 to 23.67 and PixelGen from 11.81 to 7.53, while PixelGen without REPA already outperforms JiT+REPA.
Thus, perceptual supervision provides an effective objective for pixel diffusion, and it is complementary to REPA. 

\noindent\textbf{Selected DINO layer.}
\cref{exp-ablation-depth} compares DINOv2-B feature depths for the P-DINO loss.
Shallow layers at depths 6 and 9 are less effective, likely because they mainly encode low-level appearance.
The last layer at depth 12 performs best, suggesting that P-DINO benefits more from high-level semantic features than from low-level appearance cues.
Using multiple layers performs worse, likely due to conflicting supervision across feature levels.

\noindent\textbf{Threshold of the Noise-Gating Strategy.}
\cref{exp-ablation-T-clamp} studies the noise-gating threshold.
With threshold $0.0$, perceptual losses are applied at all timesteps, including high-noise timesteps where predictions are still far from clean images. This gives strong FID and precision but lower recall.
A small threshold of $0.1$ has limited effect.
Threshold $0.3$ gives a better balance by applying perceptual losses only during the last 70\% low-noise timesteps, where predicted images are more accurate.
A large threshold of $0.6$ removes too much supervision and substantially hurts FID and IS.

\section{Conclusions}
We presented PixelGen, a simple end-to-end pixel diffusion framework that augments $x$-prediction with perceptual supervision through a local LPIPS loss, a global P-DINO loss, and a noise-gating strategy. Without VAEs, PixelGen substantially narrows the gap between pixel and latent diffusion. PixelGen surpasses strong latent baselines on ImageNet in the no-CFG setting, and on text-to-image generation it reaches a GenEval score of 0.79 with only 6 days of training on 8$\times$H800 GPUs.

PixelGen is still inferior to the strongest latent baseline under CFG and relies on pretrained perceptual encoders. Future work includes designing better pixel-space samplers and CFG strategies, and incorporating richer perceptual objectives such as adversarial losses to further improve performance.

\bibliographystyle{unsrtnat}
\bibliography{references}

\newpage
\appendix
\onecolumn

\section{Empirical Analysis of Noise Gating}
\label{appendix:noise_gating}
\begin{figure}[h]
    \centering
    \includegraphics[width=0.98\linewidth]{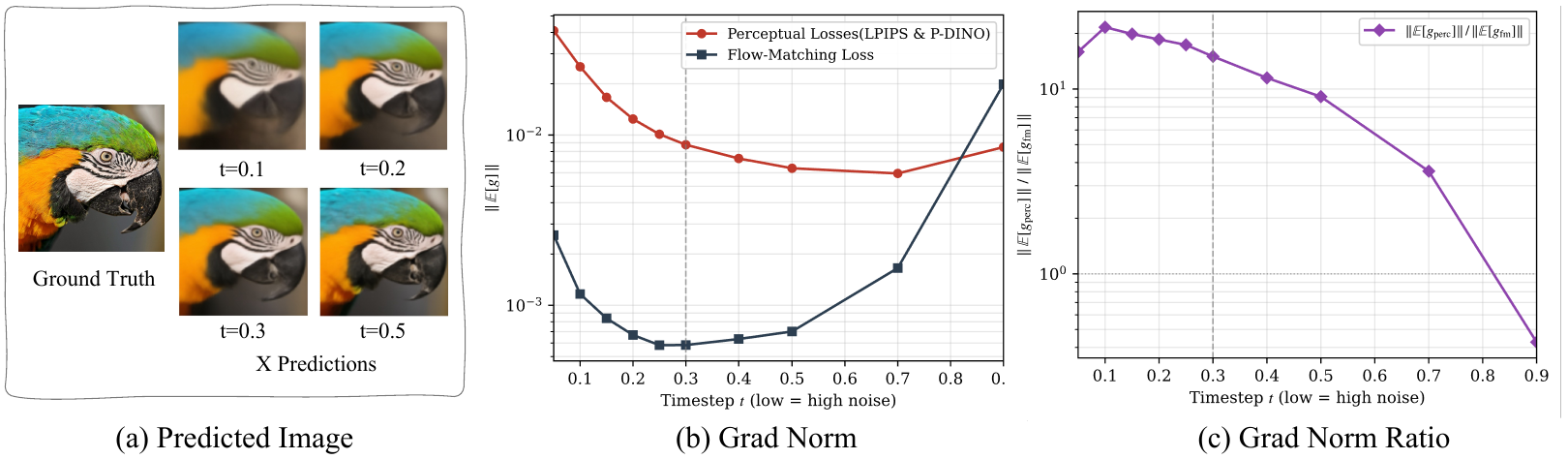}
    \caption{Empirical analysis of perceptual gradients across timesteps. (a) Predicted clean images $x_\theta$ at different timesteps. At high-noise timesteps (e.g., $t=0.1, 0.2$), the predictions are blurry and lack fine details. (b) Mean gradient norm of the perceptual losses on $x_\theta$ across different $t$. (c) Ratio between the gradient norm of the perceptual losses and that of the flow matching loss.}
    \label{fig:appendix_vis_norm}
\end{figure}

This section provides additional empirical observations that support the noise-gating design in \cref{sec:noise_gating}. Following the convention in \cref{sec:overview}, $t=0$ corresponds to pure noise and $t=1$ to clean data, so small $t$ denotes high-noise timesteps. The statistics in \cref{fig:appendix_vis_norm}(b,c) are computed on the pretrained JiT baseline in \cref{tab-baseline-comparison} with $50{,}000$ ImageNet samples. Concretely, we sweep $t$ over a uniform grid in $[0,1]$. For each $t$, we sample a clean image $x$ and Gaussian noise $\epsilon\sim\mathcal{N}(0,I)$, build the noisy input $x_t=tx+(1-t)\epsilon$, and obtain the prediction $x_\theta=\operatorname{net}_\theta(x_t,t,c)$. We then compute the gradient of each loss with respect to $x_\theta$ and report its $\ell_2$ norm averaged over the $50{,}000$ samples.

\noindent\textbf{Predicted images at high-noise timesteps.}
\cref{fig:appendix_vis_norm}(a) visualizes $x_\theta$ across timesteps. At high-noise timesteps such as $t=0.1$ and $t=0.2$, the predicted images are noticeably blurry and miss fine details. Directly aligning such predictions with the clean ground-truth image in a perceptual feature space requires the model to match details that are difficult to recover from highly noisy inputs, which may push samples away from the natural image distribution and reduce sample coverage.

\noindent\textbf{Magnitude of perceptual gradients.}
\cref{fig:appendix_vis_norm}(b) reports the mean gradient norm of the perceptual losses on $x_\theta$ as a function of $t$, computed as described above. The gradient norm is noticeably larger at high-noise timesteps (e.g., $t<0.3$) than at low-noise timesteps. Combined with \cref{fig:appendix_vis_norm}(a), this suggests that the comparatively large perceptual gradients at high-noise timesteps are partly driven by the large gap between the blurry $x_\theta$ and the clean ground-truth, and may therefore be less reliable than at low-noise timesteps.

\noindent\textbf{Relative scale to the flow matching gradient.}
\cref{fig:appendix_vis_norm}(c) plots the ratio between the gradient norm of the perceptual losses and that of the flow matching loss on $x_\theta$. The ratio is also larger at high-noise timesteps, indicating that the perceptual term tends to contribute a relatively larger share of the overall gradient there. Since $x_\theta$ is still far from a clean image in this regime, such gradients may over-emphasize clean-image feature matching during early denoising. It may reduce the sample coverage of generated images.

\section{More Implementation Details}
\subsection{Baseline Comparisons}
\label{appendix: baseline_train_details}
In this subsection, we summarize the settings used for all baseline comparisons.
In the baseline comparisons, all diffusion models are trained on ImageNet at 256$\times$256 resolution for 200k iterations using a large DiT variant. Following previous works~\cite{dit,wang2025pixnerd}, we use a global batch size of 256 and the AdamW optimizer with a constant learning rate of 1e-4. 
Both baseline and PixelGen adopt SwiGLU~\cite{llama2, llama1}, RoPE2d~\cite{rope}, and RMSNorm, and are trained with lognorm sampling and REPA~\cite{repa}.
The patch size of DiT's input is set to 16 for both baseline and our PixelGen.
The only modification on the baseline is to add two complementary perceptual losses. 
For inference, we use 50 Euler steps without classifier-free guidance~\cite{cfg} (CFG) for all models except PixelFlow~\cite{pixelflow}, which requires 100 steps. The timeshift is set to 1.0 for all experiments in \cref{tab-baseline-comparison}.

\subsection{Class-to-Image Generation}
\label{appendix: c2i_train_details}
This subsection describes additional implementation details for class-to-image generation.
The batch size and learning rate follow the default settings previously described. We use a global batch size of 256 and the AdamW optimizer with a constant learning rate of 1e-4. The time sampler uses logit-normal distribution over $t$: $\text{logit}(t){\sim}\mathcal{N}(-0.8, 0.8^2)$, which aligns with JiT~\cite{li2025jit}. We train the PixelGen-XL for 160 epochs, and use an autograd operation to balance gradients between flow-matching loss and perceptual losses after 80 epochs.
We set the CFG scale to 2.25. The guidance interval~\cite{interval_guidance} is set to (0.1, 0.9). For evaluation, we use a Heun sampler with 50 inference steps following JiT~\cite{li2025jit}. The timeshift is set to 2.0 to match the time sampler.

\subsection{Text-to-Image Generation}
\label{appendix: t2i_train_details}
We adopt Qwen3-1.7B~\cite{yang2025qwen3} as the text encoder. To improve the alignment of frozen text features \cite{fluid}, we jointly train several transformer layers on the frozen text features similar to Fluid~\cite{fluid}. The total batch size is 1536 for $256\times256$ resolution pretraining and 512 for $512\times512$ resolution pretraining. Following PixNerd~\cite{wang2025pixnerd}, we pretrain PixelGen on $256\times256$ resolution for 200K steps and pretrain on $512\times512$ resolution for 80K steps. We further fine-tune the pretrained PixelGen on BLIP3o-60k with 40k steps at the $512\times512$ resolution following PixNerd. We adopt gradient clipping to stabilize training. \textit{The whole training only takes about 6 days on 8$\times$ H800 GPUs.} We use the Adams-2nd solver with 25 steps as the default choice for sampling. The CFG scale is set to 4.0. We leave the native resolution~\cite{nit} or native aspect training~\cite{seedream2, seedream3, mogao} as future work.

\subsection{Experiment Configurations}
Table \ref{appendix:tab_config} summarizes the experiment configurations for PixelGen-L/16, PixelGen-XL/16, and PixelGen-XXL/16.
In practice, we follow the training setups from previous works such as DiT~\cite{dit}, SiT~\cite{sit}, and PixNerd~\cite{wang2025pixnerd}. 

\begin{table}[t]
\centering
\small
\caption{{Configurations of Experiments.}}
\resizebox{1.0\width}{!}{
\begin{tabular}{l |ccc}
\toprule
 & \textbf{PixelGen-L} & \textbf{PixelGen-XL} & \textbf{PixelGen-XXL} \\
\shline
\rowcolor[gray]{0.9}\multicolumn{4}{l}{\textbf{architecture}} \\
DiT depth & 22 & 28 & 16  \\
hidden dim & 1024 & 1152 & 1536  \\
heads & 16 & 16 & 24  \\
params & 459M & 676M & 1.1B  \\
bottleneck dim & 128 & - & 256\\
dropout & 0.0 & 0.1 & 0.0 \\
image size &256 & 256 & 512 \\
patch size & &16 &\\
\midline
\rowcolor[gray]{0.9}\multicolumn{4}{l}{\textbf{training}} \\
optimizer & \multicolumn{3}{c}{AdamW, $\beta_1, \beta_2=0.9, 0.999$} \\
batch size & 256 & 256 & 1536/512 \\
learning rate & \multicolumn{3}{c}{1e-4} \\ 
lr schedule & \multicolumn{3}{c}{constant} \\
weight decay & \multicolumn{3}{c}{0} \\ 
ema decay & \multicolumn{3}{c}{0.9999} \\
time sampler & \multicolumn{3}{c}{$\text{logit}(t){\sim}\mathcal{N}(\mu, \sigma^2)$, $\mu$ = -0.8, $\sigma$ = 0.8 } \\
noise scale & \multicolumn{3}{c}{1.0} \\
clip of (1-t) & \multicolumn{3}{c}{0.05} \\
\midline
\rowcolor[gray]{0.9}\multicolumn{4}{l}{\textbf{sampling}} \\
ODE solver & Euler & Heun & Adams-2nd \\
ODE steps & 50 & 50 & 25 \\
time steps & \multicolumn{3}{c}{linear in [0, 1.0]} \\
timeshift & 1.0 & 2.0 & 3.0 \\
CFG scale & - & 2.25 & 4.0 \\
CFG interval  & \multicolumn{3}{c}{{[0.1, 0.9] (if used)}} \\
\bottomrule
\end{tabular}
}
\label{appendix:tab_config}
\end{table}

\section{Text-to-Image Prompts}
\label{appendix:textual_prompts}

Below, we list the prompts used for text-to-image generation in \cref{fig:t2i_visualization}.
These prompts cover a mix of animals, people, and scenes to evaluate semantic understanding and visual detail generation.

\begin{minipage}{1.0\columnwidth}\vspace{0mm}    \centering
\begin{tcolorbox} 
    \centering
\begin{itemize}
\setlength{\itemsep}{2pt}
    
    \item A baby cat stands on two legs, wearing a chothes.
    \item A kungfu panda is wielding a sword in realistic style.
    \item A fox sleeping inside a large transparent lightbulb.
    \item An extremely happy American Cocker Spaniel is smiling and looking up at the camera with his head tilted to one side.
    \item  A man sipping coffee on a sunny balcony filled with potted plants, wearing linen clothes and sunglasses, basking in the morning light.
    \item A beautiful girl with hair flowing like a cascading waterfall.
    \item Dreamlike portrait with soft neon glow and painterly textures.
    \item A raccoon wearing a detective's hat, observing something with a magnifying glass.
    \item Close-up of an aged man with weathered features and sharp blue eyes peering wisely from beneath a tweed flat cap.
\end{itemize}
\end{tcolorbox}
\end{minipage}

\section{Pseudocode for PixelGen}
\definecolor{codeblue}{rgb}{0.25,0.5,0.5}
\definecolor{codekw}{rgb}{0.85, 0.18, 0.50}

\definecolor{codesign}{RGB}{0, 0, 255}
\definecolor{codefunc}{rgb}{0.85, 0.18, 0.50}
\definecolor{commentcolor}{rgb}{0.25,0.5,0.5}

\lstdefinelanguage{PythonFuncColor}{
  language=Python,
  keywordstyle=\color{black}\bfseries,
  commentstyle=\color{codeblue},
  stringstyle=\color{orange},
  showstringspaces=false,
  basicstyle=\ttfamily\small,
  literate=
    {*}{{\color{codesign}* }}{1}
    {-}{{\color{codesign}- }}{1}
    {+}{{\color{codesign}+ }}{1}
    {/}{{\color{codesign}/ }}{1}
    {dataloader}{{\color{codefunc}dataloader}}{1}
    {sample_t}{{\color{codefunc}sample\_t}}{1}
    {randn}{{\color{codefunc}randn}}{1}
    {randn_like}{{\color{codefunc}randn\_like}}{1}
    {jvp}{{\color{codefunc}jvp}}{1}
    {stopgrad}{{\color{codefunc}stopgrad}}{1}
    {l2_loss}{{\color{codefunc}l2\_loss}}{1}
    {LPIPS}{{\color{codefunc}LPIPS}}{1}
    {P-DINO}{{\color{codefunc}P-DINO}}{1}
    {kmeans}{{\color{codefunc}kmeans}}{1}
    {colormap}{{\color{codefunc}colormap}}{1}
    {plot}{{\color{codefunc}plot}}{1}
    {ZigZagIndices}{{\color{codefunc}ZigZagIndices}}{1}
    {DCT2D}{{\color{codefunc}DCT2D}}{1}
    {Sum}{{\color{codefunc}Sum}}{1}
    {max}{{\color{codefunc}max}}{1}
    {mean}{{\color{codefunc}mean}}{1}
    {len}{{\color{codefunc}len}}{1}
    {RGB2YCbCr}{{\color{codefunc}RGB2YCbCr}}{1}
    {patchify}{{\color{codefunc}patchify}}{1}
}

\lstset{
  language=PythonFuncColor,
  backgroundcolor=\color{white},
  basicstyle=\fontsize{8pt}{9.5pt}\ttfamily\selectfont,
  columns=fullflexible,
  breaklines=true,
  captionpos=b,
  mathescape=true 
}

\begin{algorithm}[t]
\caption{Training step}
\label{alg:code_train}
\begin{lstlisting}
# $\color{commentcolor} net_\theta$: DiT network
# $\color{commentcolor} x$: training batch
# $\color{commentcolor} c$: class label or textual prompt

$t$ = sample_t()
$\epsilon$ = randn_like($x$)

$x_t$ = $t$ * $x$ + $(1-t)$ * $\epsilon$
$v$ = ($x$ - $x_t$) / (1 - $t$)

$x_\theta$ = $net_\theta$($x_t$, $t$, $c$)
$v_\theta$ = ($x_\theta$ - $x_t$) / (1 - $t$)
$g_t$ = ($t \ge \tau$)

loss$_\text{FM}$ = l2_loss($v_\theta$ - $v$)
loss$_\text{LPIPS}$ = LPIPS($x_\theta$, $x$)
loss$_\text{P-DINO}$ = P-DINO($x_\theta$, $x$)

loss = loss$_\text{FM}$ + $g_t$ * ($\lambda_1$loss$_\text{LPIPS}$ + $\lambda_2$loss$_\text{P-DINO}$) + loss$_\text{REPA}$

\end{lstlisting}
\end{algorithm}

\begin{algorithm}[t]
\caption{Sampling step}
\label{alg:code_infer}
\begin{lstlisting}
# $\color{commentcolor} x_t$: current samples at t

$x_\theta$ = $net_\theta$($x_t$, $t$, $c$)
$v_\theta$ = ($x_\theta$ - $x_t$) / (1 - $t$)

$x_{t\_{next}}$ = $x_t$ + ($t\_next$ - $t$) * $v_{\theta}$


\end{lstlisting}
\end{algorithm}

In \cref{alg:code_train}, we provide the pseudocode for the training step of PixelGen. PixelGen follows the pipeline of JiT~\cite{li2025jit} and additionally introduces two complementary perceptual losses on the predicted image $x_\theta$.
A REPA~\cite{repa} loss is used in both our Baseline and PixelGen.
\cref{alg:code_infer} provides the pseudocode for the sampling steps.

\section{More Visualizations}
\begin{figure*}[t]
    \centering
    \includegraphics[width=0.95\linewidth]{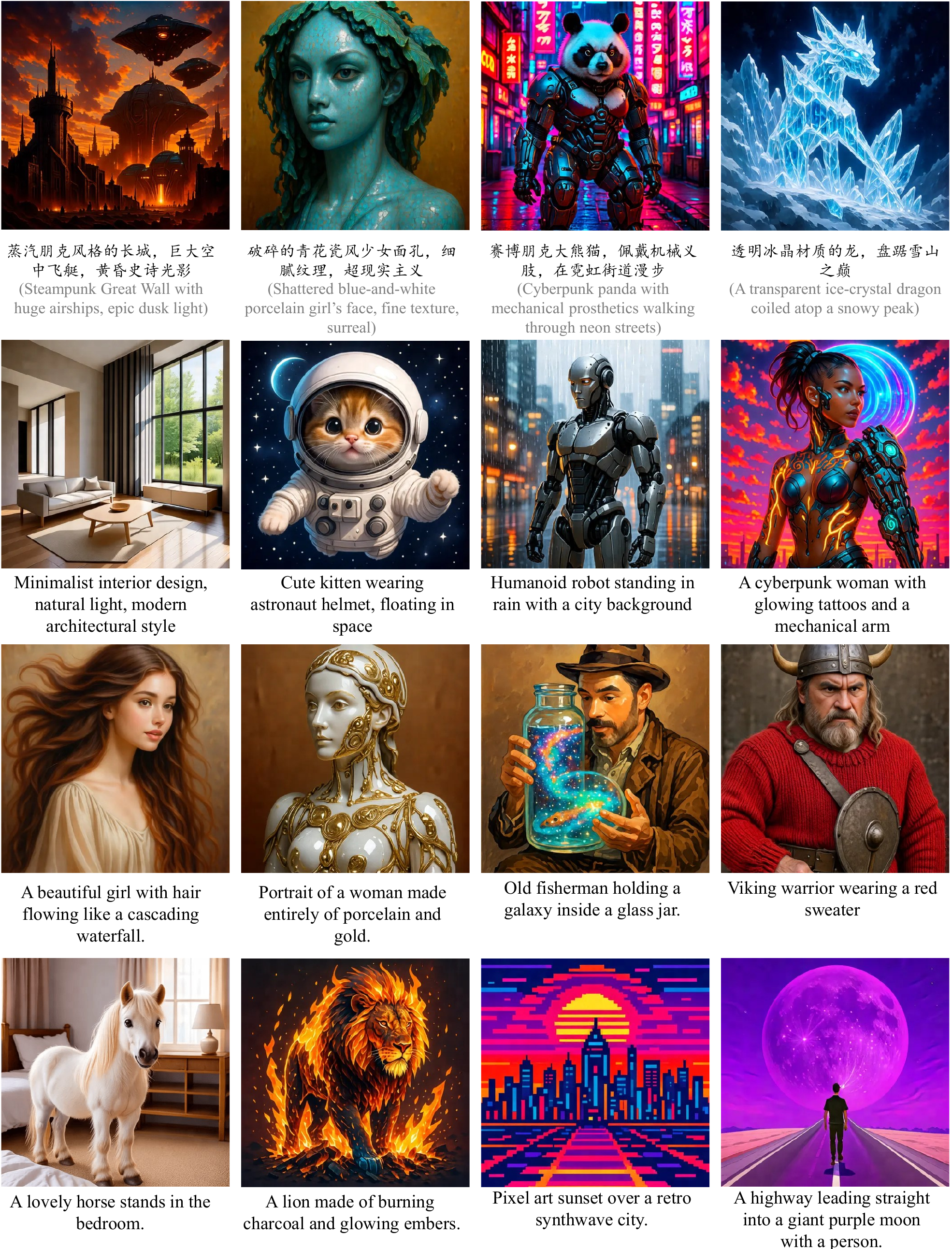}
    \caption{More qualitative results of text-to-image generation at a 512$\times$512 resolution. Our PixelGen supports multiple languages with the Qwen3 text encoder, such as Chinese and English.}
    \label{fig:visualization_512_t2i}
\end{figure*}

\begin{figure*}[t]
    \centering
    \includegraphics[width=0.95\linewidth]{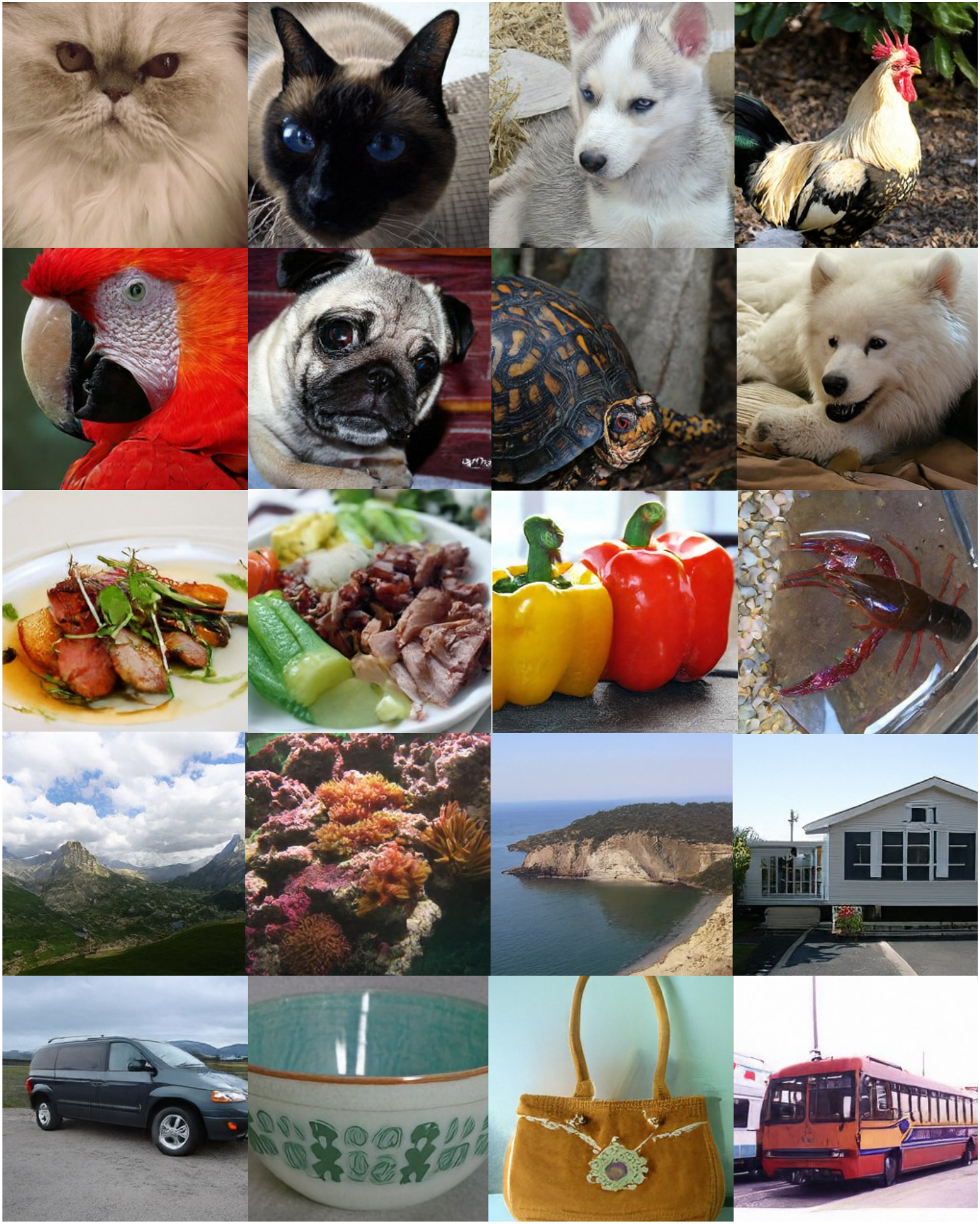}
    \caption{More qualitative results of class-to-image generation at a 256$\times$256 resolution with CFG. The CFG scale is set to 3.0.}
    \label{fig:visualization_256_imagenet}
\end{figure*}

In this section, we provide more visualizations, including text-to-image generation in \cref{fig:visualization_512_t2i}, and class-to-image generation at a 256$\times$256 resolution in \cref{fig:visualization_256_imagenet}. Our PixelGen supports multiple languages with the Qwen3 text encoder after pretraining on the BLIP3o dataset~\cite{blip3o}, such as Chinese and English.

\clearpage
\newpage
\section*{NeurIPS Paper Checklist}

\begin{enumerate}

\item {\bf Claims}
    \item[] Question: Do the main claims made in the abstract and introduction accurately reflect the paper's contributions and scope?
    \item[] Answer: \answerYes{}.
    \item[] Justification: The abstract and \cref{sec:intro} state the main contributions, including perceptual supervision for pixel diffusion, the method details, and the ImageNet and GenEval results. The experimental sections report the corresponding quantitative results.

\item {\bf Limitations}
    \item[] Question: Does the paper discuss the limitations of the work performed by the authors?
    \item[] Answer: \answerYes{}.
    \item[] Justification: The conclusion includes a limitations and future work paragraph noting the remaining gap to the strongest latent baseline under CFG and the reliance on frozen perceptual encoders. \Cref{sec-exp-class2img} also discusses the CFG-related performance gap and leaves improved samplers and CFG strategies for future work.

\item {\bf Theory assumptions and proofs}
    \item[] Question: For each theoretical result, does the paper provide the full set of assumptions and a complete (and correct) proof?
    \item[] Answer: \answerNA{}.
    \item[] Justification: The paper is empirical and methodological, and does not introduce formal theoretical results or theorem/proof statements.

    \item {\bf Experimental result reproducibility}
    \item[] Question: Does the paper fully disclose all the information needed to reproduce the main experimental results of the paper to the extent that it affects the main claims and/or conclusions of the paper (regardless of whether the code and data are provided or not)?
    \item[] Answer: \answerYes{}.
    \item[] Justification: \Cref{sec:method,sec-exp-baseline,sec-exp-class2img,sec-exp-text2img} describe the model objectives, losses, datasets, optimization settings, sampling procedures, and evaluation metrics. The appendix further provides baseline settings, class-to-image and text-to-image training details, experiment configurations, and pseudocode for training and inference.

\item {\bf Open access to data and code}
    \item[] Question: Does the paper provide open access to the data and code, with sufficient instructions to faithfully reproduce the main experimental results, as described in supplemental material?
    \item[] Answer: \answerYes{}.
    \item[] Justification: We use open-access evaluation benchmarks such as ImageNet and GenEval, and provide the text-to-image prompts used for qualitative experiments in the appendix. We are organizing our code and reproduction instructions, and plan to open-source both soon.

\item {\bf Experimental setting/details}
    \item[] Question: Does the paper specify all the training and test details (e.g., data splits, hyperparameters, how they were chosen, type of optimizer) necessary to understand the results?
    \item[] Answer: \answerYes{}.
    \item[] Justification: The experiments section specifies the datasets, metrics, backbone sizes, optimizers, batch sizes, learning rates, training durations, CFG settings, samplers, and inference steps. Additional hyperparameters and model configurations are summarized in the appendix.

\item {\bf Experiment statistical significance}
    \item[] Question: Does the paper report error bars suitably and correctly defined or other appropriate information about the statistical significance of the experiments?
    \item[] Answer: \answerNo{}.
    \item[] Justification: The paper reports standard generation metrics such as FID, Inception Score, precision/recall, and GenEval, but does not include error bars or statistical significance tests. Multiple full training runs would be computationally expensive at the reported model and dataset scales.

\item {\bf Experiments compute resources}
    \item[] Question: For each experiment, does the paper provide sufficient information on the computer resources (type of compute workers, memory, time of execution) needed to reproduce the experiments?
    \item[] Answer: \answerYes{}.
    \item[] Justification: \Cref{tab-baseline-comparison} reports training speed, wall-clock hours, memory, and inference time for the baseline comparison, and \cref{sec-exp-baseline,sec-exp-text2img} report the use of 8$\times$H800 GPUs and the 6-day text-to-image training cost. The appendix also lists the relevant training and sampling configurations.
    
\item {\bf Code of ethics}
    \item[] Question: Does the research conducted in the paper conform, in every respect, with the NeurIPS Code of Ethics \url{https://neurips.cc/public/EthicsGuidelines}?
    \item[] Answer: \answerYes{}.
    \item[] Justification: To the best of our knowledge, the research conforms to the NeurIPS Code of Ethics. The work uses standard image-generation benchmarks and publicly described datasets/models, and does not involve human-subject experiments.

\item {\bf Broader impacts}
    \item[] Question: Does the paper discuss both potential positive societal impacts and negative societal impacts of the work performed?
    \item[] Answer: \answerYes{}.
    \item[] Justification: The paper discusses that PixelGen can efficiently train a text-to-image model in only 6 days on 8$\times$H800 GPUs, which may make high-quality generative modeling more accessible and computationally efficient.
    
\item {\bf Safeguards}
    \item[] Question: Does the paper describe safeguards that have been put in place for responsible release of data or models that have a high risk for misuse (e.g., pre-trained language models, image generators, or scraped datasets)?
    \item[] Answer: \answerNA{}.
    \item[] Justification: We do not release any data or models. We do not believe that the algorithm for pixel diffusion has a high risk for misuse

\item {\bf Licenses for existing assets}
    \item[] Question: Are the creators or original owners of assets (e.g., code, data, models), used in the paper, properly credited and are the license and terms of use explicitly mentioned and properly respected?
    \item[] Answer: \answerYes{}.
    \item[] Justification:  All existing models and datasets used in this work are properly credited with appropriate references, and their licenses and terms of use have been fully respected.

\item {\bf New assets}
    \item[] Question: Are new assets introduced in the paper well documented and is the documentation provided alongside the assets?
    \item[] Answer: \answerNA{}.
    \item[] Justification: The paper does not currently introduce or release a new dataset, codebase, or model checkpoint as a new asset.

\item {\bf Crowdsourcing and research with human subjects}
    \item[] Question: For crowdsourcing experiments and research with human subjects, does the paper include the full text of instructions given to participants and screenshots, if applicable, as well as details about compensation (if any)? 
    \item[] Answer: \answerNA{}.
    \item[] Justification: The paper does not involve crowdsourcing experiments or research with human subjects.

\item {\bf Institutional review board (IRB) approvals or equivalent for research with human subjects}
    \item[] Question: Does the paper describe potential risks incurred by study participants, whether such risks were disclosed to the subjects, and whether Institutional Review Board (IRB) approvals (or an equivalent approval/review based on the requirements of your country or institution) were obtained?
    \item[] Answer: \answerNA{}.
    \item[] Justification: The paper does not involve crowdsourcing or human-subject research, so IRB approval or equivalent review is not applicable.

\item {\bf Declaration of LLM usage}
    \item[] Question: Does the paper describe the usage of LLMs if it is an important, original, or non-standard component of the core methods in this research? Note that if the LLM is used only for writing, editing, or formatting purposes and does \emph{not} impact the core methodology, scientific rigor, or originality of the research, declaration is not required.
    \item[] Answer: \answerYes{}.
    \item[] Justification: The text-to-image setup in \cref{sec-exp-text2img} and the appendix explicitly state that Qwen3-1.7B is used as the text encoder, with additional transformer layers trained on frozen text features. This LLM component is part of the text-to-image system rather than merely a writing or formatting aid.

\end{enumerate}

\end{document}